\def\year{2018}\relax
\documentclass[letterpaper]{article} 
\usepackage{aaai18}  
\usepackage{times}  
\usepackage{helvet}  
\usepackage{courier}  
\usepackage{url}  
\usepackage{graphicx}  
\usepackage{todonotes}
\begin{document}
%
\title{A Unified Implicit Dialog Framework for Conversational Search}

\author{
Song Feng, R. Chulaka Gunasekara, Sunil Shashidhara, Kshitij P. Fadnis and Lazaros C. Polymenakos\\
IBM Thomas J. Watson Research Center\\
Yorktown Heights, NY, USA 10598\\
}

\maketitle

\begin{abstract}
We propose a unified \textit{Implicit} Dialog framework for goal-oriented, information seeking tasks of Conversational Search applications. It aims to enable dialog interactions with domain data without replying on explicitly encoded the rules but utilizing the underlying data representation to build the components required for dialog interaction, which we refer as \textit{Implicit} Dialog in this work. The proposed framework consists of a pipeline of End-to-End trainable modules. A centralized knowledge representation is used to semantically ground multiple dialog modules. An associated set of tools are integrated with the framework to gather end users' input for continuous improvement of the system.  
The goal is to facilitate development of conversational systems by identifying the components and the data that can be adapted and reused across many end-user applications. 
We demonstrate our approach by creating conversational agents for several independent domains.
\end{abstract}

\section{Introduction}
The demands for interacting with commercial services through chat applications, a.k.a, Conversational Search (CS), are increasing rapidly. The goal is to provide end users with information from the underlying knowledge base of a domain  (e.g., car insurance, pizza ordering, apparel shopping). 
Conventional dialog system development typically starts by explicitly modeling the possible queries of users and then build the components that will allow conversational interaction. By contrast, the implicit dialog approach starts from the application's data representation, and then builds the related dialog components. Thus, the interaction of the user with the system is not explicitly coded in rules, but it is done implicitly based on task's data representation. 

However, just querying a knowledge base can become very challenging. 
Applications typically have numerous hand-crafted query rules that are often fused with machine learning components trained specifically for a single domain. This entangling of generic and domain specific knowledge makes it nearly impossible to adapt an existing CS application to new domains. 
To address these challenges, we focus on methodologies that facilitate dialog interaction with specific types of knowledge representation, and propose a unified development framework for goal-oriented, information seeking conversation systems.
We facilitate interactive search, enabling the developers to identify and share common building blocks across various applications in a domain-agnostic manner. 

We design our framework assuming that a domain knowledge base is available, and can be obtained from the corresponding commercial website. Example schema of such websites are illustrated at \url{schema.org}. The combination of the domain knowledge base, the permitted queries to the knowledge base, and the application logic 
allows the creation of dialog interaction. In particular, we first scan the knowledge base and build a central knowledge representation that can semantically ground multiple dialog subtasks, such as intent labeling, state tracking, and issuing API calls to the domain database. The resulting framework is end-to-end modular and trainable, in line with recently proposed approaches for building conversational agents \cite{rasa}, \cite{truong2017maca}. Those frameworks typically require application developers to code their own dialog logic and provide the training data for dialog subtasks \cite{rasa}. However, both the design of dialog management, and annotating the chat data are costly. Our framework is integrated with a feedback module that allows the conversational agent to take advantage of the expertise of a wide variety of stakeholders: end users,  domain experts, and human annotators. The module is initialized with the central knowledge representation, and collects targeted feedback that can be directly consumed by the learning modules 
for continuous improvement. 

We demonstrate the proposed framework on two different domains: apartment renting in NYC and restaurant finding.

\section{System Overview}
\begin{figure}[!ht]
    \centering
    \scalebox{0.85}{
    \includegraphics[width=0.5\textwidth]{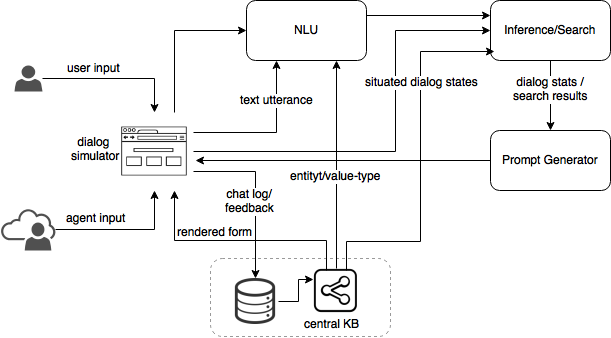}}
    \caption{Architecture of the Framework}
    \label{fig:system}
\end{figure}

As depicted in Figure \ref{fig:system}, the proposed framework includes several core modules such as Natural Language Understanding (intent labeler), Inference (state tracker and query generator), a prompt generator and a dialog environment simulator for data collections. The input of these components are all initialized by the central knowledge representation.

\subsection{Central Knowledge Representation} 
We introduce a Central Knowledge Representation (CKR), generated based on the domain knowledge base.  
The CKR covers the domain entities (e.g., ``\texttt{apartment}''), their semantic relations (e.g., ``\texttt{has-attribute}''),
and generic characteristics (e.g., expected data types, ranges, operations) associated with the entities, all generated in a domain agnostic way. We then generate the dialog components based on the extracted CKR, thus allowing seamless adaptation from one domain to another. 

\subsection{State Tracking and Inference}
To identify the user intents we use a natural language classifier \cite{yang2016hierarchical} trained on chat data\footnote{ \url{https://datasets.maluuba.com/Frames}.}, where the domain dependent tokens are delexicalised. To facilitate interactive search across domains for CS, we first identify a set of generic operations, such as {\texttt{ADD, DELETE, UPDATE}} that are useful for updating the dialog states and forming queries. 
The best performing classifier shows a promising accuracy of $89\%$ in a test dataset containing over $200$ utterances. 

Dialog state is tracked using semantic matching. We map user utterance to the elements in CKR in three consecutive steps, i.e, (1) literal matching, (2) fuzzy matching - which identifies the approximate matches between entities, and (3) vector representation matching - which supports matching word embedding representations of related entities.  
Our framework is modular and supports plug-n-play of different semantic matching methodologies.

Given the last user intent, dialog state and previous dialog acts, the conversational agent either issue an API call using the query graph generated based on CKR or request more information to optimize the search experience. Information theory based measures are used to determine the slot to be requested.

\begin{figure}[!ht]
    \centering
    \includegraphics[width=0.45\textwidth]{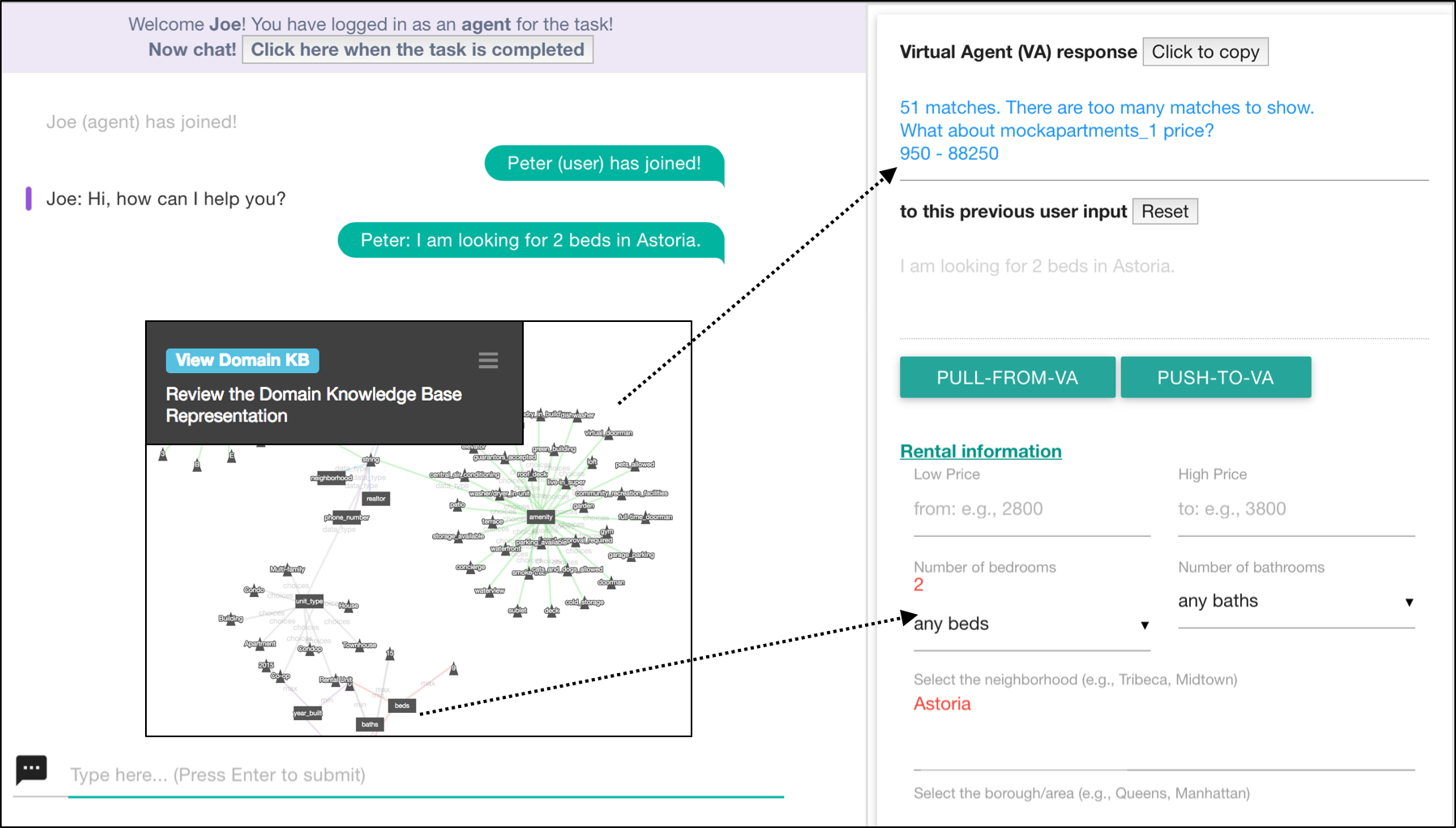}
    \caption{Dialog Simulator Interface}
    \label{fig:woz}
\end{figure}

\subsection{Dialog Simulator}
The dialog simulator generates various conversational environments in a Wizard-of-OZ fashion \cite{schatzmann2005quantitative}, which are configured to interface with the crowd-sourcing platforms. Figure \ref{fig:woz} presents the interface rendered based on relevant content in CKR. It can simulate and log the process followed by human agents: a) to access the requested information based on identified user intent, and b) to convert that information to dialog prompts. The human agents can also provide feedback on the system's output during the real-time interactions. Compared to many existing chat log annotating tools, our framework captures  real-time information that is difficult for human annotators to provide offline. 

\section{Case Studies}
We apply the proposed framework on two independent domains: (1) rental apartments (2) restaurants. For the former, we create a mock database that includes information about \texttt{\#bedroom}, \texttt{transportation}, \texttt{location}. As illustrated in Figure \ref{fig:woz}, the conversational agent could list search results based on user’s request and provide the guidance on how to optimize the query toward the goal. For the development of the later, we only replace the domain database and re-generate CKR. The conversational agent appears to perform comparably on the new domain.

\section{Conclusion and Future Work}
We propose a unified framework of Implicit Dialog for Conversational Search grounded by a centralized knowledge representation. It facilitates fast prototyping and aims at making existing development and chat data reusable and adaptable to new domains. As future work, we plan to conform the central knowledge representation with \url{schema.org} to enable the broader practice on domain adaptions.

\bibliographystyle{aaai}
\bibliography{bib}

\end{document}